\documentclass[a4paper, 10pt, conference]{ieeeconf} 
\IEEEoverridecommandlockouts 
\usepackage{graphicx} 
\usepackage{amsmath} 
\usepackage{amssymb}  
\usepackage[hidelinks]{hyperref}
\usepackage{bm}
\usepackage{subcaption}
\usepackage{xcolor}

\title{\LARGE \bf Improving Low-Cost Teleoperation: Augmenting GELLO with Force*}

\author{Shivakanth Sujit$^{\dagger,1}$, Luca Nunziante$^{\dagger,1}$, Dan Ogawa Lillrank$^{1}$,\\
Rousslan Fernand Julien Dossa$^{1}$ and Kai Arulkumaran$^{1}$
\thanks{*This work was supported by JST under Moonshot R\&D Grant Number JPMJMS2012.}
\thanks{$^{\dagger}$These authors contributed equally.}%
\thanks{$^{1}$Araya Inc., Tokyo, Japan\quad {\tt\small sujit\_shivakanth@araya.org}}%
}

\begin{document}

\maketitle
\thispagestyle{empty}
\pagestyle{empty}

\begin{abstract}
In this work we extend the low-cost GELLO teleoperation system, initially designed for joint position control, with additional force information. Our first extension is to implement force feedback, allowing users to feel resistance when interacting with the environment. Our second extension is to add force information into the data collection process and training of imitation learning models. We validate our additions by implementing these on a GELLO system with a Franka Panda arm as the follower robot, performing a user study, and comparing the performance of policies trained with and without force information on a range of simulated and real dexterous manipulation tasks. Qualitatively, users with robotics experience preferred our controller, and the addition of force inputs improved task success on the majority of tasks.
\end{abstract}

\section{INTRODUCTION}

In the last few years, there has been a rapid increase in the scope of abilities demonstrated by robots, driven by advances in machine learning (ML). Examples of such abilities include champion-level drone racing \cite{kaufmann2023champion} and quadruped parkour \cite{zhuang2023robot}, achieved through reinforcement learning (RL), or wheeled/humanoid loco-manipulation \cite{fu2024mobile,seo2023deep}, achieved through imitation learning (IL). Although RL has proved powerful for learning controllers to accomplish specific tasks, IL has become popular in multi-task settings, as RL reward functions need to be engineered for each task under consideration. Even with as simple an IL algorithm as behavioural cloning (BC) \cite{pomerleau1988alvinn}, researchers have been able to deploy mobile manipulators to perform tasks in office settings, bootstrapped by teleoperation data \cite{brohan2022rt,brohan2023rt}.

Beyond purely autonomous behaviour, human teleoperation of robots can provide a range of useful outcomes. The most widely-deployed use-case is in robotic surgery \cite{howe1999robotics}, though use-cases can vary from engineering/maintenance in remote locations \cite{luo2020combined} to providing social interaction and employment for disabled people \cite{takeuchi2020avatar}. Although teleoperation can apply to locomotion or other tasks, manipulation stands to benefit the most from improvements in teleoperation, as manually designing controllers or formulating RL reward functions for dexterous manipulation is challenging and difficult to  generalise \cite{s23073762}. Therefore, we focus the rest of our exposition on teleoperation for manipulation tasks.

The development of better teleoperation systems has benefits for both direct human control of robots, as well as for data collection for IL-based control policies. Devices such as gamepads or the SpaceMouse\footnote{\url{https://3dconnexion.com/us/spacemouse/}.} have been commonly used for teleoperation in the past, and nowadays with widening public adoption, virtual reality controllers have also become more popular for robot teleoperation \cite{zhang2018deep}. However, these devices have few degrees of freedom (DoF), and hence are limited to end-effector control in task/Cartesian space. The downside of control in task space is under-specification of the robot's joint positions, and encountering singularities that prevent motion.

This can be obviated through leader-follower teleoperation systems, where the joints of a leader robot can be directly manipulated by a human, and the follower robot mimics this movement. Unfortunately, leader-follower systems typically assume two of the same robot, and for many robots cost (or even space) make this solution prohibitive.

\begin{figure}[t]
  \centering
  \includegraphics[width=\columnwidth]{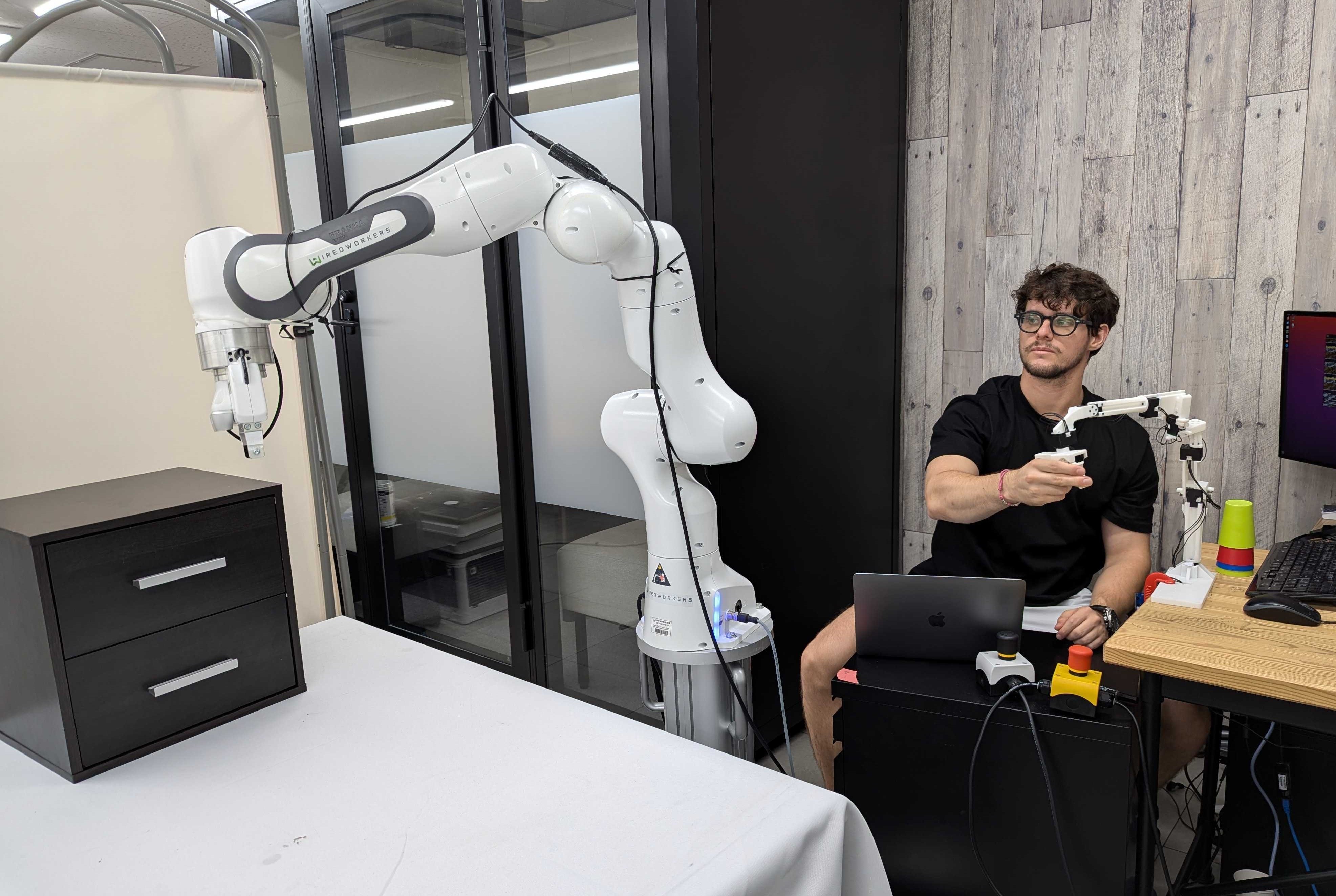}
  \caption{The low-cost GELLO teleoperation system (right), controlling a Franka Panda 7 DoF robot arm (centre).}
  \label{fig:overview}
  \vspace{-10pt}
\end{figure}

This limitation is overcome by the GELLO system \cite{wu2023gello}, which uses a low-cost (less than \$300) kinematically-equivalent 3D-printed ``robot''. The authors built equivalents for Franka, UR5, and xArm robots, with the robotics community subsequently contributing further models.\footnote{\url{https://github.com/wuphilipp/gello_mechanical}.} With open source hardware and software, GELLO significantly lowers the barriers to entry for leader-follower teleoperation.

However, the original GELLO system does not provide any feedback to the user, and ignores force information. In this work we extend the control scheme to enable force-position control, i.e., force feedback, so that users can feel interactions with the environment (Fig.~\ref{fig:overview}). We test our system with a user study; although there was no significant difference in usability across the population, those experienced with robots preferred having the feedback. In addition, we use data collected from our augmented controller to train an action chunking Transformer (ACT) policy \cite{zhao2023learning} on 4 manipulation tasks in simulation and the real world, and we find that performance was improved in 3/4 of the tasks.

\section{METHOD}
\subsection{Leader-Follower Teleoperation}
In leader-follower teleoperated systems, we can identify three main components: a leader device, in contact with the human operator; a follower device, that interacts with the environment; and a communication channel between the two systems. Both the leader and follower have their own dedicated controllers. Controllers may differ in terms of information exchanged over the channel, sensors used, and feedback provided to the user, but in teleoperation there are two common objectives that they can aim to accomplish \cite{lawrence1993stability}:
\begin{itemize}
    \item Stability: we desire values of the state of the two systems (position, velocity) to be bounded in response to bounded external inputs (forces applied by the operator).
    \item Transparency: the human operator should feel a direct interaction with the remote environment during the execution of the task. In free space, without collision, the resistance felt by the human user should be low and the follower should accurately track the desired trajectory. When an interaction between the follower and the environment happens, the stiffness perceived by the human operator should be the same as the one experienced on the follower side.
\end{itemize}
Developing haptic interfaces for position and force feedback, and adopting a leader device with dexterity and a kinematic structure as similar as possible to the follower device, are therefore key to designing a suitable teleoperation user interface and guaranteeing high levels of transparency.

It is also possible to use kinematically-dissimilar leader and follower robots, but these need to be coupled at the end effector level. For example, this is often the case for large industrial or surgical robots that are controlled by smaller haptic interfaces. However, this setup means that the human operator is made oblivious to kinematic constraints on the follower side, such as reaching joint limits, self collisions, and configuration singularities. 

GELLO overcomes the downsides of end-effector mapping as it is kinematically equivalent to the leader robot. It is also cost- and space-effective, using open-sourced 3D-printed parts and off-the-shelf servomotors. Overall, it is one of the most accessible and user-friendly teleoperation interfaces available today \cite{rea2022still}.

\subsection{Force Feedback}

In general, both force and position information can be exchanged between the leader and follower systems, and different control systems vary in which kind of information they share. Although it is possible to perform teleoperation without feedback to the user, as in the original GELLO, it has been shown in prior work that providing haptic feedback improves task performance \cite{wildenbeest2012impact,triantafyllidis2020study,nitsch2012meta}. The main controllers used in teleoperated systems that can provide haptic feedback are: position-position (PP), force-position (FP), and four channel (4C) \cite{aliaga2004experimental}:
\begin{itemize}
    \item PP: the only information shared between the robots is their position. The leader position acts as a reference for the follower, who will try to track it, e.g., by means of a proportional–derivative controller. Similarly, the leader will receive and track the follower position. In this scheme, no force information is exchanged and force reflection emerges as a result of a growing tracking error on the follower side when in contact with the environment. 
    \item FP: the follower robot acts identically as in the PP controller, tracking the position of the leader. However, when the follower comes in contact with the environment, the resulting forces are (scaled and) transmitted to the leader side so that the operator can feel them. In this force reflection scheme, the magnitude of the scaling term is a critical factor on system stability due to the system's asymmetry \cite{penin1997design}. 
    \item 4C: this control scheme, a.k.a. bilateral control, not only achieves position tracking, but also uses the principle of action and reaction in the force transmission. The follower replicates the leader's motion, and the leader exerts feedback forces equal to the interaction forces on the follower from the environment \cite{sabanovic2011motion}. The position and force responses of the leader and follower are identical, resulting in ``ideal kinesthetic coupling'' \cite{yokokohji1994bilateral}.
\end{itemize}

4C is the control scheme that achieves the best performance in terms of tracking and force reflection, and has been used in recent works in teleoperation for IL \cite{sasagawa2020imitation,sakaino2022imitation,buamanee2024bi}. The common characteristic in these works is that they use identical robots on the leader and on the follower side, enforcing the action-reaction relation between forces at the joint level.

Unfortunately, 4C requires force sensors/estimation on both the leader and the follower side. The GELLO leader robot is a kinematically-scaled version of the follower, but no dynamic relation (e.g., relative mass) is given. Moreover, GELLO is equipped with motors that fail to sustain its own weight. This last aspect hinders any dynamic parameter identification procedure needed for force estimation. This therefore precludes implementing 4C control for GELLO.

\begin{figure}[t]
  \centering
  \includegraphics[width=\linewidth]{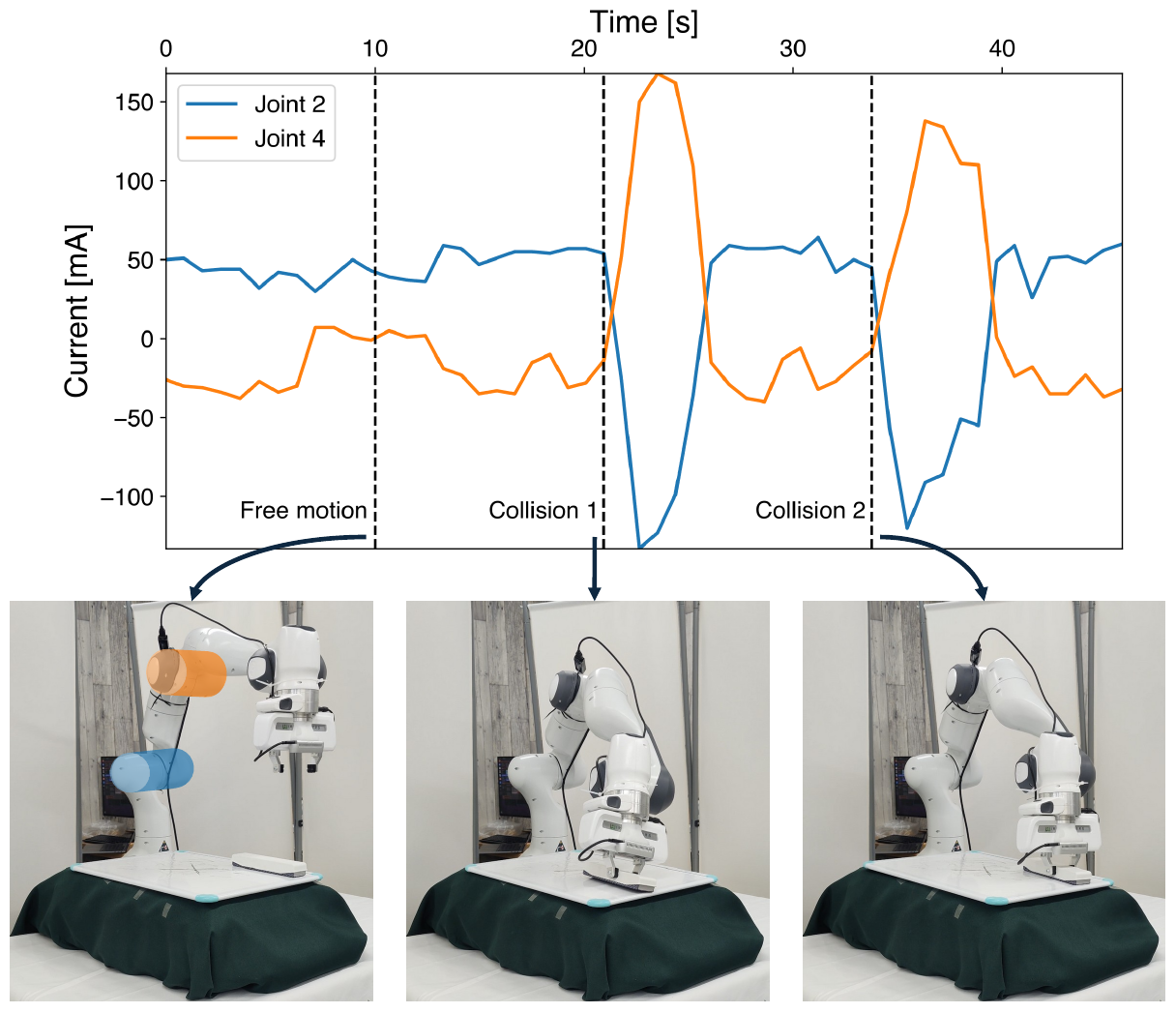}
  \caption{Current feedback for joints two (blue) and four (orange) during teleoperation for the whiteboard erasing task, highlighting three key instants: free motion, collision 1, and collision 2. These joints exhibit the highest activity during collisions, with sharp current spikes indicating the system's response to external forces. During the free motion phase, non-zero currents are observed due to noise in the internal computation of external torque on the Franka arm.}
  \label{fig:contact}
  \vspace{-15pt}
\end{figure}

However, we can exploit the kinematic equivalence to adopt the FP control scheme. Given an external wrench, $\bm{\mathcal{F}}_{ext} \in \mathbb{R}^6$, applied by the follower end-effector on the environment, the corresponding joint torque is
\begin{equation}
    \bm{\tau}^{ext}_f = \bm{J}^\top_f \bm{\mathcal{F}}_{ext},
    \label{eq:jac_transp}
\end{equation}
where ${\bm{J}_f \in \mathbb{R}^{6 \times n}}$ is the follower end-effector Jacobian, and $n$ is the number of joints. Since the leader's Denavit-Hartenberg parameters are a scaled version of the follower's, the linear part of the leader's end-effector Jacobian is a scaled version of the corresponding linear part in $\bm{J}_f$, while the angular part stays unchanged. Thus, the component of $\bm{\tau}^{ext}_f$ due to the external linear force should be scaled down when reflected on the leader's side, while the component coming from the external moment should be transmitted as is. In practice, as the linear contribution is dominant in our experiments, and this breakdown would require an online pseudo-inversion of $\bm{J}^\top_f$, the joint torque we impose on the leader is simply
\begin{equation}
    \bm{\tau}^{ref}_l = -k_f \bm{\tau}^{ext}_f,
    \label{eq:gello_ctrl}
\end{equation}
where $k_f > 0 $ is a scalar gain (scaling factor) and $\bm{\tau}^{ref}_l \in \mathbb{R}^n$ is the reference torque for the leader. As the external torque $\bm{\tau}^{ext}_f$ is non-zero only when the follower robot interacts with the environment, during free motion the leader will not have any force feedback. As depicted in Fig.~\ref{fig:contact}, the user will feel resistance only when there is an environment interaction on the follower side.

The value of the force-feedback ratio $k_f$ in \eqref{eq:gello_ctrl} plays a crucial role in the system stability and transparency. There is an upper limit on its value in order to have a stable system, and higher values also show an oscillatory response in the force reflection, masking the benefit of such feedback. A detailed analysis can be found in \cite{penin1997design}.

On the follower side, we use the leader's joint positions as reference for a joint space impedance controller \cite{tsetserukou2009isora}. We also experimented with providing a joint velocity reference to the leader, but the resulting motion on the leader side turned out to be less smooth and stable.

\subsection{Imitation Learning}
\begin{figure*}[t]
  \centering
  \includegraphics[width=0.95\textwidth]{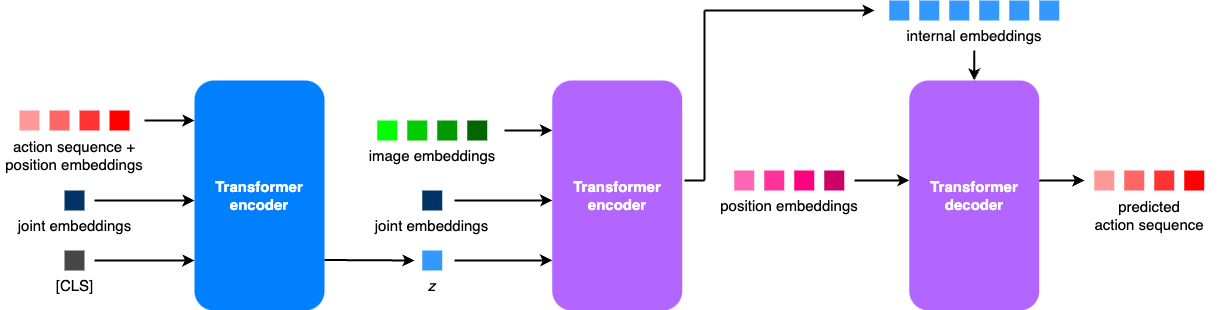}
  \caption{The ACT architecture. During deployment, the Transformer encoder (middle) takes in images, joint information, and the latent vector $z$ as input, and produces embeddings which are used by the Transformer decoder to predict actions. During training, an additional Transformer encoder (left) is used to train the entire set of models as a CVAE.}
  \label{fig:act}
\end{figure*}

\begin{figure*}
  \centering
  \begin{subfigure}[b]{0.24\textwidth}
    \centering
    \includegraphics[width=\textwidth]{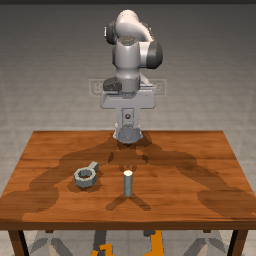}
    \caption{Nut Assembly}
    \label{fig:sim_insertion}
  \end{subfigure}
  \begin{subfigure}[b]{0.24\textwidth}
    \centering
    \includegraphics[width=\textwidth]{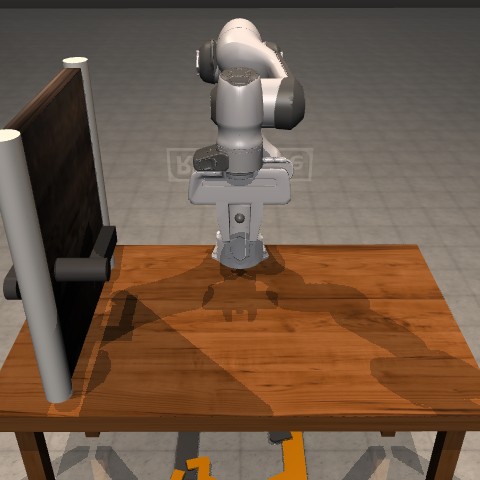}
    \caption{Door Opening}
    \label{fig:sim_opening}
  \end{subfigure}
  \begin{subfigure}[b]{0.24\textwidth}
    \centering
    \includegraphics[width=\textwidth]{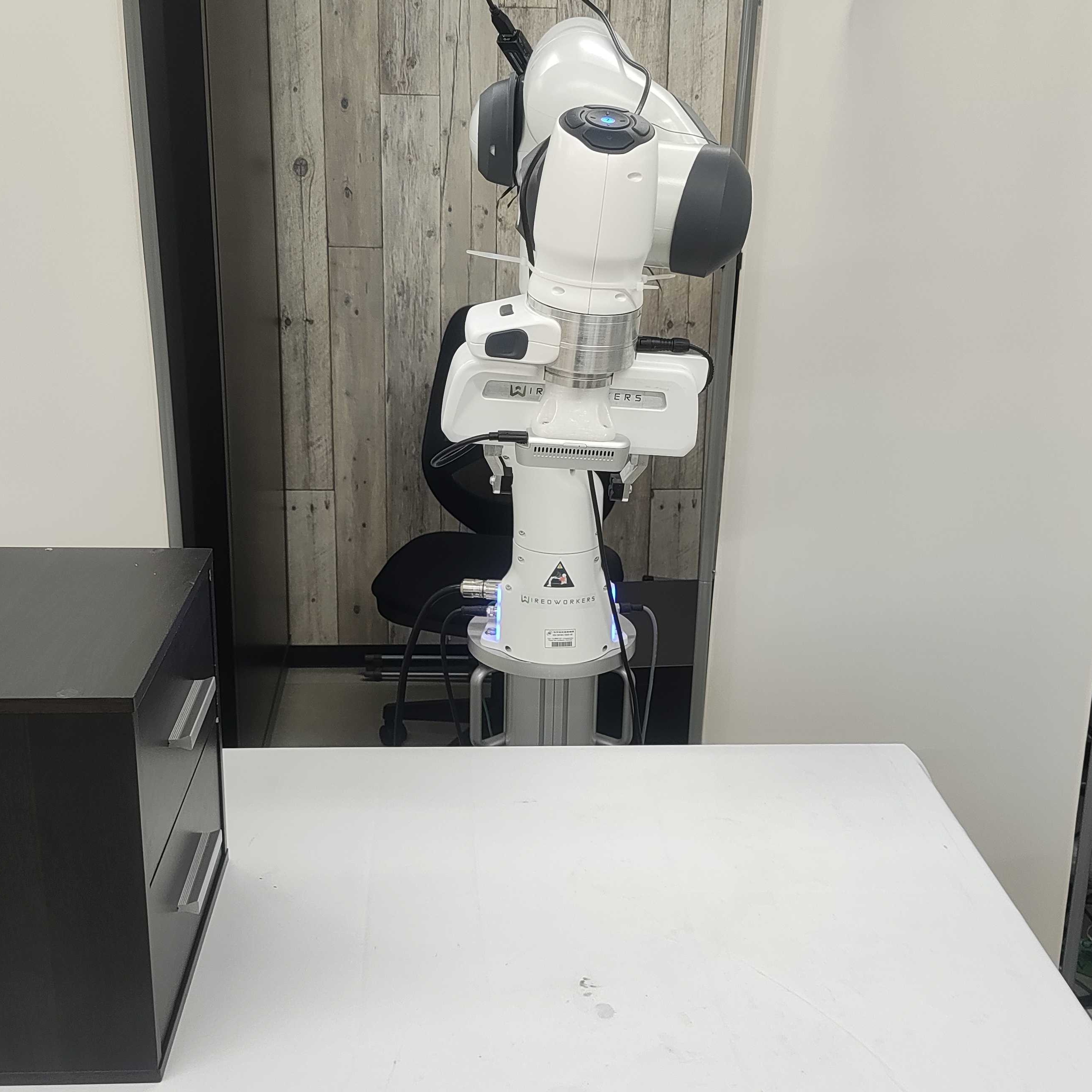}
    \caption{Drawer Opening}
    \label{fig:real_draw}
  \end{subfigure}
  \begin{subfigure}[b]{0.24\textwidth}
    \centering
    \includegraphics[width=\textwidth]{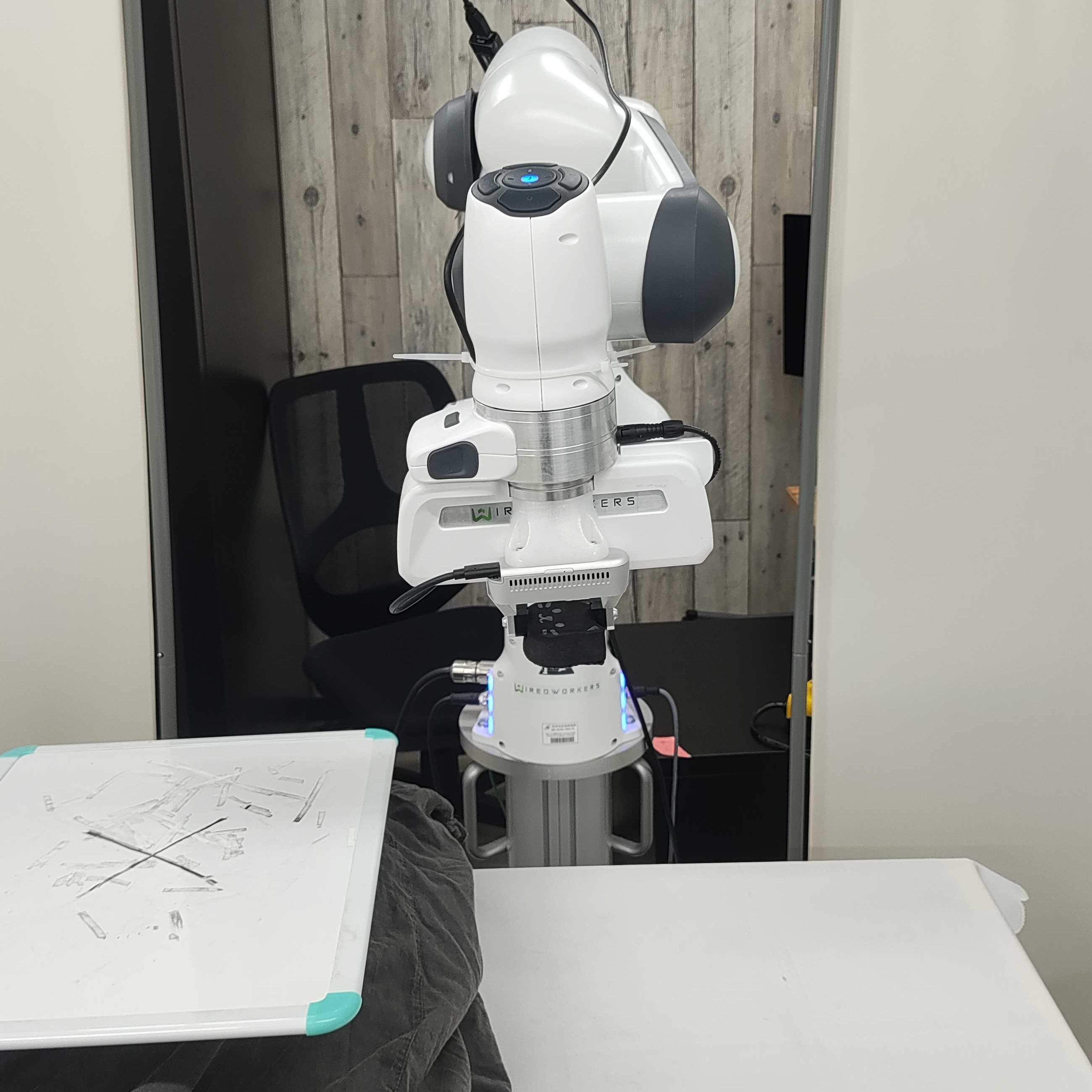}
    \caption{Whiteboard Erasing}
    \label{fig:real_erase}
  \end{subfigure}
  \caption{Set of manipulation tasks for IL with different action types: grasping, placing, pulling, and wiping.}
  \label{fig:tasks}
  \vspace{-10pt}
\end{figure*}

By collecting data through teleoperation, nowadays it is possible to train IL policies that can accomplish challenging manipulation tasks. The simplest IL method is BC \cite{pomerleau1988alvinn}, which is the application of supervised learning to expert trajectories. In general, with a control policy $\pi$, BC maximises the likelihood of expert actions, $a$, at the states, $s$, encountered by the expert, at timesteps $t$: $\max \pi(a_t|s_t)$.

One of the recent revolutions in the application of IL in robotics is the development of expressive policy formulations: in particular, ACT \cite{zhao2023learning} and diffusion policies \cite{chi2023diffusionpolicy}. By using powerful models to predict sequences of actions (``action chunks''), these methods are able to better capture multimodal action distributions and temporal dependencies between actions. These models predict sequences of $k$ actions: $\pi(a_{t:t+k}|s_t)$.

We base our method on ACT (Fig.~\ref{fig:act}), which trains a Transformer model \cite{vaswani2017attention} to predict action sequences using BC. During deployment, a Transformer encoder takes in the current state (camera images, embedded via convolutional neural networks, and the joint information, embedded via a linear layer), and a latent variable, $z$ (set to the zero vector during deployment), and produces a set of embeddings using self-attention. These are then used as keys and values for cross-attention in a Transformer decoder that takes position embeddings as input (for queries), and produces the action sequence as output. ACT is trained as a conditional variational autoencoder (CVAE) \cite{sohn2015learning}, with an additional Transformer encoder that takes in the joint information (embedded via a linear layer), the action sequence (plus position embeddings), and a [CLS] token\footnote{An additional token that can be transformed to create the output \cite{devlin2018bert}.}, and produces the latent variable $z$ that is taken as input by the rest of the ACT model that is used during deployment. In addition to maximising the likelihood of the action sequence, $z$ is regularised by minimising the Kullback-Leibler divergence between sampled $z$ vectors and a Normal distribution: $D_\text{KL}(z \Vert \mathcal{N}(0, I))$.

For deployment, the authors of ACT predicted action sequences at every time step, and averaged the predictions over time to produce smoother trajectories (as opposed to following the predicted action sequences open-loop). We also use this temporal averaging for deployment.

Whilst ACT originally only used position data, Bi-ACT \cite{buamanee2024bi} and Comp-ACT \cite{kamijo2024learning} have extended it to use force information. Bi-ACT implements 4C (bilateral) control, which is made possible by using the same leader and follower robots (ROBOTIS OpenMANIPULATOR-X). Comp-ACT uses forward dynamics compliance control \cite{scherzinger2017forward}, which is an impedance control scheme in Cartesian space, and hence retains the disadvantages of end-effector mapping. We also show the benefits of incorporating force information in IL with ACT, but with the low-cost GELLO teleoperation system for joint control.


\section{EXPERIMENTS}

For all of our experiments, in both simulation and the real world, we use a 7 DoF Franka Emika Panda robot arm. Communication with the robot is managed via the Franka Control Interface, which provides access to information on the robot's state, including estimates of the external torque $\bm{\tau}^{ext}$ required to implement \eqref{eq:gello_ctrl}. The leader robot, GELLO, is equipped with DYNAMIXEL motors that support current control, enabling the implementation of \eqref{eq:gello_ctrl} by applying an additional conversion factor.\footnote{Discussion with the GELLO authors revealed that they observed motor disconnections when they tried to implement feedback, due to torque or power supply limitations. Based on this, we used a more powerful power supply and scaled down the reflected force.} The Franka robot is also equipped with one Intel RealSense D435i RGBD camera mounted on the wrist, and one Intel RealSense D435i RGBD overlooks the scene. Our simulation tasks are built in RoboHive \cite{kumar2024robohive}, with MuJoCo \cite{todorov2012mujoco} as the physics simulator. We used a control frequency and data collection rate of 50Hz.

For IL, we use ACT, with its default hyperparameters. The model has an embedding layer of size 512, and linear layers of size 3200 in the Transformers. The KL regularisation weight $\beta$ is set to 50. We predict action chunks of size 50.

\subsection{Tasks}

Many manipulation tasks may be achievable with position and visual information, but can be improved through the inclusion of force information. We constructed 4 tasks (Fig.~\ref{fig:tasks}), 2 in simulation, and 2 in the real world, to examine this:
\begin{enumerate}
    \item Nut Assembly (Sim). A version of the Nut Assembly task from robosuite \cite{zhu2020robosuite}, which involves picking up a circular nut and placing it on a peg. There is a small tolerance, requiring precision for placement. To reduce slippage, we slightly increased the length of the handle on the nut. The nut position and orientation is randomised every rollout (for both demonstrations and evaluation). We consider it a success when the nut is placed so that the peg is completely on the inside.
    \item Door Opening (Sim). A version of the Door Opening task from robosuite, which involves opening a door with a revolute joint by turning the handle and pulling the door open. The position and orientation of the door is randomised every rollout. We consider it a success if the door angle reaches at least 0.3 rad.
    \item Drawer Opening (Real). This task involves opening the top drawer of a chest of drawers, where the drawer has a prismatic joint. Although it should be possible to open with motion in a straight line, factors such as the end-effector angle and static and dynamic friction make this more challenging than it appears at first. We randomised the position of the drawer by up to 6cm for every rollout. We consider it a success if the drawer is opened by over 15cm.
    \item Whiteboard Erasing (Real). This task involves using an eraser to remove marks on a whiteboard, with the eraser already placed in the gripper. For the user study, users had to erase 2 crosses on the whiteboard that were drawn by the experimenter. For IL experiments, the experimenter drew different marks, providing natural variation, in the centre of the board, and collected demonstrations of erasing these. We consider it a success if the majority of the mark is erased.
\end{enumerate}

\subsection{User Study}
We recruited a total of 20 volunteers for a user study (6 female, average age of 32.5 years, standard deviation of 7.7 years) to evaluate our force feedback controller, vs. the default controller without feedback. Users were briefed on the teleoperation setup and shown an example of the erasing task with the real robot. If they consented, we then proceeded with the study. Our study was given ethical approval by the Shiba Palace Clinic Ethics Review Committee.

We told users that they would be evaluating 2 different teleoperation controllers, but they were not informed of the differences. Users were then given 2 minutes to experiment with and practice the task with the first controller, before being asked to erase the crosses on the whiteboard. This procedure was then repeated for the second controller. Users were finally asked to fill in a NASA task load index (TLX) survey \cite{hart1988development} for controllers 1 and 2, as a measure of perceived workload. To avoid confounding from learning effects, the order of presenting the force feedback or original GELLO controller were randomised for each user.

\begin{table}[t]
  \caption{NASA TLX survey results (mean $\pm$ 1 standard deviation; lower values are better). Across all 6 items, there were no significant differences on perceived workload between the two controllers.}
  \vspace{-5pt}
  \label{tbl:survey}
  \begin{center}
    \begin{tabular}{|c||c|c|}
      \hline
      & No Feedback & Force Feedback \\
      \hline
      Mental Demand   & $33.0 \pm 25.7$ & $37.2 \pm 25.5$ \\
      Physical Demand & $38.2 \pm 22.1$ & $40.2 \pm 19.8$ \\
      Temporal Demand & $22.5 \pm 21.5$ & $30.8 \pm 26.8$ \\
      Performance     & $20.5 \pm 25.6$ & $20.5 \pm 20.5$ \\
      Effort          & $43.0 \pm 25.1$ & $48.2 \pm 22.9$ \\
      Frustration     & $29.2 \pm 21.5$ & $35.5 \pm 20.1$ \\
      \hline
    \end{tabular}
  \end{center}
  \vspace{-15pt}
\end{table}

The results from the NASA TLX survey are presented in Table \ref{tbl:survey}. Considering the TLX scores as ordinal data, we used the Wilcoxon signed-rank test for paired non-parametric data to test if there was a significant difference between the two conditions. The largest difference between conditions was on physical demand (W-statistic: 19.0, p-value: 0.06), but this is not statistically different at a significance level of 0.05, applying the Bonferroni correction for multiple statistical tests. Therefore, there was no significant perceptual difference to users between the two controllers.

We also gathered qualitative feedback from users. Most users had no experience with robots or teleoperation, and, if they noticed a difference between controllers, equated the force feedback with ``resistance'' to their control. Conversely, the few users who were familiar with robots recognised that one controller had force feedback and the other did not. These users preferred the controller with force feedback, claiming that it provided complementary information to visual feedback, allowing operators to confirm that their actions align with their intentions. This indicates that we can expect force feedback to help teleoperators with experience.

\subsection{Imitation Learning}
We used the open source ACT code for training policies, modifying the input to the policy to contain leader joint positions and joint torque due to external forces, $\bm{\tau}^{ext}_f$. The prediction targets for ACT were the GELLO joint positions. We collected 25 demonstrations per task, with 22 demonstrations used for training, and 3 reserved for validation. We trained policies using 3 different random seeds per task, and evaluated each policy 15 times, for a total of 45 evaluation rollouts per task and input type. Following the authors' recommendations \cite{zhao2023learning}, all policies were trained for 5000 epochs, even though the validation loss plateaued earlier. 

\begin{table}[t]
  \caption{Success rate on the 4 tasks using different inputs. Using force inputs increased performance across most tasks.}
  \vspace{-5pt}
  \label{tbl:example}
  \begin{center}
    \resizebox{\columnwidth}{!}{%
    \begin{tabular}{|c||c|c|c|c|}
      \hline
      Inputs           & Nut             & Door & Drawer       & Whiteboard\\
      \hline
      Position         & $\mathbf{0.60} \pm 0.13$   & $0.96 \pm 0.21$ & $0.62 \pm 0.48$ & $0.24 \pm 0.43$\\
      Position + Force & $0.42 \pm 0.10$ & $\mathbf{1.0} \pm 0.0$ & $\mathbf{0.93} \pm 0.25$ & $\mathbf{0.36} \pm 0.48$\\
      \hline
    \end{tabular}%
    }
  \end{center}
  \vspace{-15pt}
\end{table}

As reported in Table \ref{tbl:example}, including force information improved performance on most tasks. The most noticeable improvement was in the Drawer Opening task, where all failures were due to failing to grasp the drawer handle well. With force information, the policy was able to recognise missed grasps by not feeling a resistance when pulling away from the drawers, and autonomously recovered by going back and performing a correct grasp. Without force information, the policy was unable to recover. In contrast, with Door Opening, successful manipulation of the handle is much clearer for both policies.

For Whiteboard Erasing, the overall low success rate is mostly due to occlusions: once the robot approaches the scribble whilst holding the eraser, the scribble is not seen anymore by either of the two cameras.

Nut Assembly is the sole task where arguably force sensitivity is not important, and in this task the position-only policy performs better. However, most failures for both policies were in picking up the nut initially, so we believe that the performance of both policies could be improved by collecting more demonstrations.

Performance across seeds was relatively consistent, i.e., for a given task each seed had approximately the same success rate. This means that ACT is robust to random initialisation, which is particularly useful in robotics applications.

\section{CONCLUSIONS}
In this study we improved upon an existing teleoperation system, GELLO, by including force feedback for users, and force inputs to improve the performance of IL on different manipulation tasks. For robots to be able to interact with uncontrolled, real world settings, where visual cues can be limiting, or even confusing, incorporating other sensor modalities such as haptic and force feedback would be essential for improved performance. We believe that our modification to the low-cost teleportation system, with an improved user experience, can help contribute to current efforts to IL, such as the collection of robotic datasets \cite{ebert2021bridge,padalkar2023open}.

\addtolength{\textheight}{-2cm} 


\section*{ACKNOWLEDGMENT}
The authors would like to thank Thanpimon Buamanee for answering questions on Bi-ACT, Alex Hattori for advice on communication with Dynamixel motors, and Marina Di Vincenzo for help with the user study.

\bibliographystyle{IEEEtran}
\bibliography{main}

\begin{thebibliography}{10}
\providecommand{\url}[1]{#1}
\csname url@rmstyle\endcsname
\providecommand{\newblock}{\relax}
\providecommand{\bibinfo}[2]{#2}
\providecommand\BIBentrySTDinterwordspacing{\spaceskip=0pt\relax}
\providecommand\BIBentryALTinterwordstretchfactor{4}
\providecommand\BIBentryALTinterwordspacing{\spaceskip=\fontdimen2\font plus
\BIBentryALTinterwordstretchfactor\fontdimen3\font minus \fontdimen4\font\relax}
\providecommand\BIBforeignlanguage[2]{{%
\expandafter\ifx\csname l@#1\endcsname\relax
\typeout{** WARNING: IEEEtran.bst: No hyphenation pattern has been}%
\typeout{** loaded for the language `#1'. Using the pattern for}%
\typeout{** the default language instead.}%
\else
\language=\csname l@#1\endcsname
\fi
#2}}

\bibitem{kaufmann2023champion}
E.~Kaufmann, L.~Bauersfeld, A.~Loquercio, M.~M{\"u}ller, V.~Koltun, and D.~Scaramuzza, ``{Champion-level Drone Racing Using Deep Reinforcement Learning},'' \emph{Nature}, vol. 620, no. 7976, pp. 982--987, 2023.

\bibitem{zhuang2023robot}
Z.~Zhuang, Z.~Fu, J.~Wang, C.~G. Atkeson, S.~Schwertfeger, C.~Finn, and H.~Zhao, ``{Robot Parkour Learning},'' in \emph{Conference on Robot Learning}.\hskip 1em plus 0.5em minus 0.4em\relax PMLR, 2023, pp. 73--92.

\bibitem{fu2024mobile}
Z.~Fu, T.~Z. Zhao, and C.~Finn, ``{Mobile ALOHA: Learning Bimanual Mobile Manipulation with Low-Cost Whole-Body Teleoperation},'' \emph{arXiv preprint arXiv:2401.02117}, 2024.

\bibitem{seo2023deep}
M.~Seo, S.~Han, K.~Sim, S.~H. Bang, C.~Gonzalez, L.~Sentis, and Y.~Zhu, ``{Deep Imitation Learning for Humanoid Loco-manipulation Through Human Teleoperation},'' in \emph{IEEE-RAS Humanoids}.\hskip 1em plus 0.5em minus 0.4em\relax IEEE, 2023, pp. 1--8.

\bibitem{pomerleau1988alvinn}
D.~A. Pomerleau, ``{ALVINN: An Autonomous Land Vehicle in a Neural Network},'' in \emph{NeurIPS}, 1988.

\bibitem{brohan2022rt}
A.~Brohan, N.~Brown, J.~Carbajal, Y.~Chebotar, J.~Dabis, C.~Finn, K.~Gopalakrishnan, K.~Hausman, A.~Herzog, J.~Hsu, \emph{et~al.}, ``{RT-1: Robotics Transformer for Real-World Control at Scale},'' \emph{arXiv preprint arXiv:2212.06817}, 2022.

\bibitem{brohan2023rt}
A.~Brohan, N.~Brown, J.~Carbajal, Y.~Chebotar, X.~Chen, K.~Choromanski, T.~Ding, D.~Driess, A.~Dubey, C.~Finn, \emph{et~al.}, ``{RT-2: Vision-Language-Action Models Transfer Web Knowledge to Robotic Control},'' \emph{arXiv preprint arXiv:2307.15818}, 2023.

\bibitem{howe1999robotics}
R.~D. Howe and Y.~Matsuoka, ``{Robotics for Surgery},'' \emph{Annual review of biomedical engineering}, vol.~1, no.~1, pp. 211--240, 1999.

\bibitem{luo2020combined}
J.~Luo, W.~He, and C.~Yang, ``{Combined Perception, Control, and Learning for Teleoperation: Key Technologies, Applications, and Challenges},'' \emph{Cognitive Computation and Systems}, vol.~2, no.~2, pp. 33--43, 2020.

\bibitem{takeuchi2020avatar}
K.~Takeuchi, Y.~Yamazaki, and K.~Yoshifuji, ``{Avatar Work: Telework for Disabled People Unable to Go Outside by Using Avatar Robots},'' in \emph{Companion of the 2020 ACM/IEEE international conference on human-robot interaction}, 2020, pp. 53--60.

\bibitem{s23073762}
\BIBentryALTinterwordspacing
D.~Han, B.~Mulyana, V.~Stankovic, and S.~Cheng, ``A survey on deep reinforcement learning algorithms for robotic manipulation,'' \emph{Sensors}, vol.~23, no.~7, 2023. [Online]. Available: \url{https://www.mdpi.com/1424-8220/23/7/3762}
\BIBentrySTDinterwordspacing

\bibitem{zhang2018deep}
T.~Zhang, Z.~McCarthy, O.~Jow, D.~Lee, X.~Chen, K.~Goldberg, and P.~Abbeel, ``{Deep Imitation Learning for Complex Manipulation Tasks from Virtual Reality Teleoperation},'' in \emph{2018 IEEE international conference on robotics and automation (ICRA)}.\hskip 1em plus 0.5em minus 0.4em\relax IEEE, 2018, pp. 5628--5635.

\bibitem{wu2023gello}
P.~Wu, Y.~Shentu, Z.~Yi, X.~Lin, and P.~Abbeel, ``{GELLO: A General, Low-Cost, and Intuitive Teleoperation Framework for Robot Manipulators},'' in \emph{Towards Generalist Robots Workshop, CoRL}, 2023.

\bibitem{zhao2023learning}
T.~Z. Zhao, V.~Kumar, S.~Levine, and C.~Finn, ``{Learning Fine-grained Bimanual Manipulation with Low-Cost Hardware},'' in \emph{RSS}, 2023.

\bibitem{lawrence1993stability}
D.~A. Lawrence, ``{Stability and Transparency in Bilateral Teleoperation},'' \emph{IEEE transactions on robotics and automation}, vol.~9, no.~5, pp. 624--637, 1993.

\bibitem{rea2022still}
D.~J. Rea and S.~H. Seo, ``{Still Not Solved: A Call for Renewed Focus on User-Centered Teleoperation Interfaces},'' \emph{Frontiers in Robotics and AI}, vol.~9, p. 704225, 2022.

\bibitem{wildenbeest2012impact}
J.~G. Wildenbeest, D.~A. Abbink, C.~J. Heemskerk, F.~C. Van Der~Helm, and H.~Boessenkool, ``The impact of haptic feedback quality on the performance of teleoperated assembly tasks,'' \emph{IEEE Transactions on Haptics}, vol.~6, no.~2, pp. 242--252, 2012.

\bibitem{triantafyllidis2020study}
E.~Triantafyllidis, C.~Mcgreavy, J.~Gu, and Z.~Li, ``Study of multimodal interfaces and the improvements on teleoperation,'' \emph{IEEE Access}, vol.~8, pp. 78\,213--78\,227, 2020.

\bibitem{nitsch2012meta}
V.~Nitsch and B.~F{\"a}rber, ``A meta-analysis of the effects of haptic interfaces on task performance with teleoperation systems,'' \emph{IEEE transactions on haptics}, vol.~6, no.~4, pp. 387--398, 2012.

\bibitem{aliaga2004experimental}
I.~Aliaga, A.~Rubio, and E.~Sanchez, ``Experimental quantitative comparison of different control architectures for master-slave teleoperation,'' \emph{IEEE transactions on control systems technology}, vol.~12, no.~1, pp. 2--11, 2004.

\bibitem{penin1997design}
L.~Pe{\~n}{\'\i}n, M.~Ferre, J.~Fernandez-Pello, R.~Aracil, and A.~Barrientos, ``Design fundamentals of master-slave systems with a force-position bilateral control scheme,'' \emph{IFAC Proceedings Volumes}, vol.~30, no.~20, pp. 605--612, 1997.

\bibitem{sabanovic2011motion}
A.~Sabanovic and K.~Ohnishi, \emph{{Motion Control Systems}}.\hskip 1em plus 0.5em minus 0.4em\relax John Wiley \& Sons, 2011.

\bibitem{yokokohji1994bilateral}
Y.~Yokokohji and T.~Yoshikawa, ``{Bilateral Control of Master-Slave Manipulators for Ideal Kinesthetic Coupling-Formulation and Experiment},'' \emph{IEEE T-RO}, vol.~10, no.~5, pp. 605--620, 1994.

\bibitem{sasagawa2020imitation}
A.~Sasagawa, K.~Fujimoto, S.~Sakaino, and T.~Tsuji, ``Imitation learning based on bilateral control for human--robot cooperation,'' \emph{IEEE Robotics and Automation Letters}, vol.~5, no.~4, pp. 6169--6176, 2020.

\bibitem{sakaino2022imitation}
S.~Sakaino, K.~Fujimoto, Y.~Saigusa, and T.~Tsuji, ``Imitation learning for variable speed contact motion for operation up to control bandwidth,'' \emph{IEEE Open Journal of the Industrial Electronics Society}, vol.~3, pp. 116--127, 2022.

\bibitem{buamanee2024bi}
T.~Buamanee, M.~Kobayashi, Y.~Uranishi, and H.~Takemura, ``{Bi-ACT: Bilateral Control-Based Imitation Learning via Action Chunking with Transformer},'' in \emph{Manipulation Skills Workshop, ICRA}, 2024.

\bibitem{tsetserukou2009isora}
D.~Tsetserukou, N.~Kawakami, and S.~Tachi, ``isora: humanoid robot arm for intelligent haptic interaction with the environment,'' \emph{Advanced Robotics}, vol.~23, no.~10, pp. 1327--1358, 2009.

\bibitem{chi2023diffusionpolicy}
C.~Chi, S.~Feng, Y.~Du, Z.~Xu, E.~Cousineau, B.~Burchfiel, and S.~Song, ``{Diffusion Policy: Visuomotor Policy Learning via Action Diffusion},'' in \emph{RSS}, 2023.

\bibitem{vaswani2017attention}
A.~Vaswani, N.~Shazeer, N.~Parmar, J.~Uszkoreit, L.~Jones, A.~N. Gomez, {\L}.~Kaiser, and I.~Polosukhin, ``{Attention is All You Need},'' in \emph{NeurIPS}, 2017.

\bibitem{sohn2015learning}
K.~Sohn, H.~Lee, and X.~Yan, ``{Learning Structured Output Representation Using Deep Conditional Generative Models},'' in \emph{NeurIPS}, 2015.

\bibitem{devlin2018bert}
J.~Devlin, ``{BERT: Pre-Training of Deep Bidirectional Transformers for Language Understanding},'' \emph{arXiv preprint arXiv:1810.04805}, 2018.

\bibitem{kamijo2024learning}
T.~Kamijo, C.~C. Beltran-Hernandez, and M.~Hamaya, ``{Learning Variable Compliance Control From a Few Demonstrations for Bimanual Robot with Haptic Feedback Teleoperation System},'' in \emph{IROS}, 2024.

\bibitem{scherzinger2017forward}
S.~Scherzinger, A.~Roennau, and R.~Dillmann, ``{Forward Dynamics Compliance Control (FDCC): A New Approach to Cartesian Compliance for Robotic Manipulators},'' in \emph{2017 IEEE/RSJ International Conference on Intelligent Robots and Systems (IROS)}.\hskip 1em plus 0.5em minus 0.4em\relax IEEE, 2017, pp. 4568--4575.

\bibitem{kumar2024robohive}
V.~Kumar, R.~Shah, G.~Zhou, V.~Moens, V.~Caggiano, A.~Gupta, and A.~Rajeswaran, ``{RoboHive: A Unified Framework for Robot Learning},'' in \emph{NeurIPS}, 2024.

\bibitem{todorov2012mujoco}
E.~Todorov, T.~Erez, and Y.~Tassa, ``{MuJoCo: A Physics Engine for Model-Based Control},'' in \emph{IROS}, 2012.

\bibitem{zhu2020robosuite}
Y.~Zhu, J.~Wong, A.~Mandlekar, R.~Mart{\'\i}n-Mart{\'\i}n, A.~Joshi, S.~Nasiriany, and Y.~Zhu, ``{robosuite: A Modular Simulation Framework and Benchmark for Robot Learning},'' \emph{arXiv preprint arXiv:2009.12293}, 2020.

\bibitem{hart1988development}
S.~G. Hart and L.~E. Staveland, ``{Development of NASA-TLX (Task Load Index): Results of Empirical and Theoretical Research},'' in \emph{Advances in psychology}.\hskip 1em plus 0.5em minus 0.4em\relax Elsevier, 1988, vol.~52, pp. 139--183.

\bibitem{ebert2021bridge}
F.~Ebert, Y.~Yang, K.~Schmeckpeper, B.~Bucher, G.~Georgakis, K.~Daniilidis, C.~Finn, and S.~Levine, ``Bridge data: Boosting generalization of robotic skills with cross-domain datasets,'' \emph{arXiv preprint arXiv:2109.13396}, 2021.

\bibitem{padalkar2023open}
A.~Padalkar, A.~Pooley, A.~Jain, A.~Bewley, A.~Herzog, A.~Irpan, A.~Khazatsky, A.~Rai, A.~Singh, A.~Brohan, \emph{et~al.}, ``Open x-embodiment: Robotic learning datasets and rt-x models,'' \emph{arXiv preprint arXiv:2310.08864}, 2023.

\end{thebibliography}

\end{document}